\documentclass[runningheads]{llncs}
\usepackage[T1]{fontenc}
\usepackage{graphicx}
\usepackage{multirow, siunitx, wrapfig, tikz, dsfont, booktabs, amsfonts, amsmath, caption, subcaption, cite}
\usepackage{cleveref} 
\usepackage{bm}
\begin{document}
\title{FlowMoDL: Model-Based Deep Learning with Conjugate-Gradient Data Consistency for Highly Accelerated 4D Flow MRI Reconstruction}
\titlerunning{FlowMoDL}
%
\author{Tristan Gottwald\inst{1,3,}$^\star$\orcidID{0009-0004-3550-9779} \and
Michelle Bruch\inst{4}\orcidID{0009-0006-2485-1741} \and
Mubashir-Ul Hassan\inst{1,2,3}\orcidID{0009-0005-1404-9176} \and
Fatma Alickovic\inst{5}\orcidID{0009-0001-6885-0269}\and
Milan Kloiber\inst{1}\orcidID{0009-0005-9204-6106}\and
Daniel Tenbrinck\inst{4}\orcidID{0000-0002-4788-9332}\and
Torsten Panholzer\inst{5}\orcidID{0000-0002-2808-6120}\and
Melanie Schaller\inst{1,3}\orcidID{0000-0002-5708-4394}\and
Jana Hutter\inst{1,2,3}\orcidID{0000-0003-3476-3500}}
\authorrunning{T. Gottwald et al.}
%
\institute{
Institute of Information Processing (tnt), Leibniz University Hannover, Germany
\and CAIMED, Leibniz University Hannover, Hannover, Germany 
\and L3S, Leibniz University Hannover, Hannover, Germany 
\and Department of Data Science, Friedrich-Alexander-Universität Erlangen-Nürnberg, Germany 
\and Institute of Medical Biostatistics, Epidemiology and Informatics, University Medical Center of the Johannes Gutenberg University Mainz, Germany \\$^\star$Corresponding author: \email{gottwald@tnt.uni-hannover.de}  }
\maketitle              
\begin{abstract}
We present FlowMoDL, an unrolled neural network for highly accelerated 4D flow MRI reconstruction that directly optimizes for both anatomical magnitude and phase-derived velocity accuracy. Building on the MoDL framework, FlowMoDL alternates a learned (3+1)D spatiotemporal denoiser with conjugate-gradient data-consistency updates based on the SENSE forward model. A novel dual-pathway conditioning scheme adapts the denoiser features and data-consistency weighting, enabling a single model to handle varying acceleration factors ($10\times$ to $50\times$). To ensure physiological accuracy, the network is trained using a deep-supervision composite loss that explicitly penalizes velocity magnitude and angular errors, stabilized by a curriculum schedule. We evaluate FlowMoDL on the multi-center CMRx4DFlow dataset against classical and deep-learning baselines (CG-SENSE, MoDL, FlowVN, and FlowMRI-Net). A key advantage of FlowMoDL is its superior gradient step efficiency. When evaluated under an equivalent, limited budget of gradient steps, competing flow-specific networks degrade significantly. In contrast, FlowMoDL robustly converges and strictly outperforms all competitors across all acceleration factors in magnitude SSIM, nRMSE, relative velocity error, and angular error, successfully recovering sharp structural details and temporally coherent velocity fields.
\end{abstract}
\section{Introduction}
Four-dimensional flow magnetic resonance imaging (4D flow MRI) captures time-resolved, volumetric anatomical and three-directional blood-flow information across the cardiac cycle~\cite{markl20124d}. However, its extensive data requirements necessitate aggressive k-space undersampling~\cite{stankovic20144d, munoz2023acceleration}. Reconstructing these highly accelerated acquisitions is challenging because the velocity field is derived entirely from relative phase differences between velocity encodings. Standard reconstruction methods prioritizing magnitude~\cite{aggarwal2018modl,heckel2024deep} are therefore insufficient, as even marginal phase errors induce severe, non-physiological velocity deviations. Furthermore, existing flow-specific deep learning architectures rely on computationally heavy recurrent mechanisms~\cite{jacobs2025flowmrinet} or neglect explicitly optimizing for phase~\cite{vishnevskiy2020flowvn}.\\
To address these limitations, we introduce FlowMoDL, a phase-aware unrolled neural network explicitly designed for highly accelerated 4D flow MRI. Built upon the MoDL framework~\cite{aggarwal2018modl}, FlowMoDL couples a (3+1)D spatiotemporal regularizer with conjugate-gradient data-consistency updates. Our primary contributions are: 1) a hybrid architecture that preserves critical inter-encoding phase relationships; 2) a dual-pathway conditioning mechanism that adapts denoiser features and data-consistency weighting to handle acceleration rates from $10\times$ to $50\times$ with a single trained model; 3) a curriculum-guided composite loss optimizing directly for flow and 4) state-of-the-art performance and gradient-step efficiency, strictly outperforming classical and specialized baselines on the multi-center CMRx4DFlow dataset.\\ The code is available at \url{https://github.com/tgottwald/FlowMoDL}.

\section{Method}
\subsection{Problem formulation}
We formulate the highly accelerated 4D flow MRI reconstruction~\cite{vishnevskiy2020flowvn,jacobs2025flowmrinet} as follows.
Given the undersampled multi-coil k-space measurements \linebreak $\mathbf{y}\in\mathbb{C}^{N_v\times N_c\times N_t\times N_z\times N_y\times N_x}$, coil sensitivity maps $S\in\mathbb{C}^{N_c\times N_z\times N_y\times N_x}$, and an undersampling mask $M\in\{0,1\}^{N_t\times N_z\times N_y}$, we aim to reconstruct the MRI volume sequence $\mathbf{x}\in\mathbb{C}^{N_v\times N_t\times N_z\times N_y\times N_x}$ and derive the corresponding 4D flow velocity field. Here, $N_v$ denotes the number of velocity encodings, $N_t$ the number of cardiac phases, $N_c$ the number of coils, and $N_z$, $N_y$, and $N_x$ the spatial dimensions. We consider $N_v=4$, where the first encoding serves as the reference encoding and the phase differences between the remaining three encodings and the reference yield the directional velocity components over the cardiac cycle~\cite{markl20124d}.
\\
Acquisition is modeled using the SENSE~\cite{pruessmann1999sense} forward operator $A$:
\begin{equation}
   A\mathbf{x} = M \mathcal{F}(S\mathbf{x}), 
\end{equation}
where $\mathcal{F}$ corresponds to the spatial Fourier transform.
The adjoint operator of $A$ is referred to as $A^{\mathsf H}$.\\
In contrast to existing 4D flow approaches based on learned multidimensional filter-bank updates~\cite{vishnevskiy2020flowvn} or recurrent complex-valued updates that propagate information across cardiac phases and reconstruction iterations~\cite{jacobs2025flowmrinet}, we adopt the MoDL framework~\cite{aggarwal2018modl} to combine a spatiotemporal learned prior with conjugate-gradient data-consistency. The reconstruction problem is formulated as
\begin{equation}
    \widehat{\mathbf{x}}
    =
    \arg\min_{\mathbf{x}}
    \frac{1}{2}
    \left\|A\mathbf{x}-\mathbf{y}\right\|_2^2
    +
    \frac{\lambda}{2}
    \left\|
    \mathbf{x}-\mathcal{D}_{\theta}(\mathbf{x})
    \right\|_2^2,
    \label{eq:reconstruction_problem}
\end{equation}
where $\mathcal{D}_{\theta}$ denotes a learned spatiotemporal denoiser and $\lambda>0$ balances measurement consistency and the learned residual prior. FlowMoDL approximates this formulation by alternating learned denoising and conjugate-gradient data-consistency updates across the unrolled cascades.\\ 
As velocity is derived from the relative phase between the directional and reference encodings, $v_i \propto \angle(\mathbf{x}_i\overline{\mathbf{x}}_0)$, reconstruction must preserve inter-encoding phase. This motivates the proposed spatiotemporal architecture and the explicit penalization of velocity magnitude and direction errors in the training objective.

\subsection{FlowMoDL}
\begin{figure}[tb]
    \includegraphics[width=\textwidth]{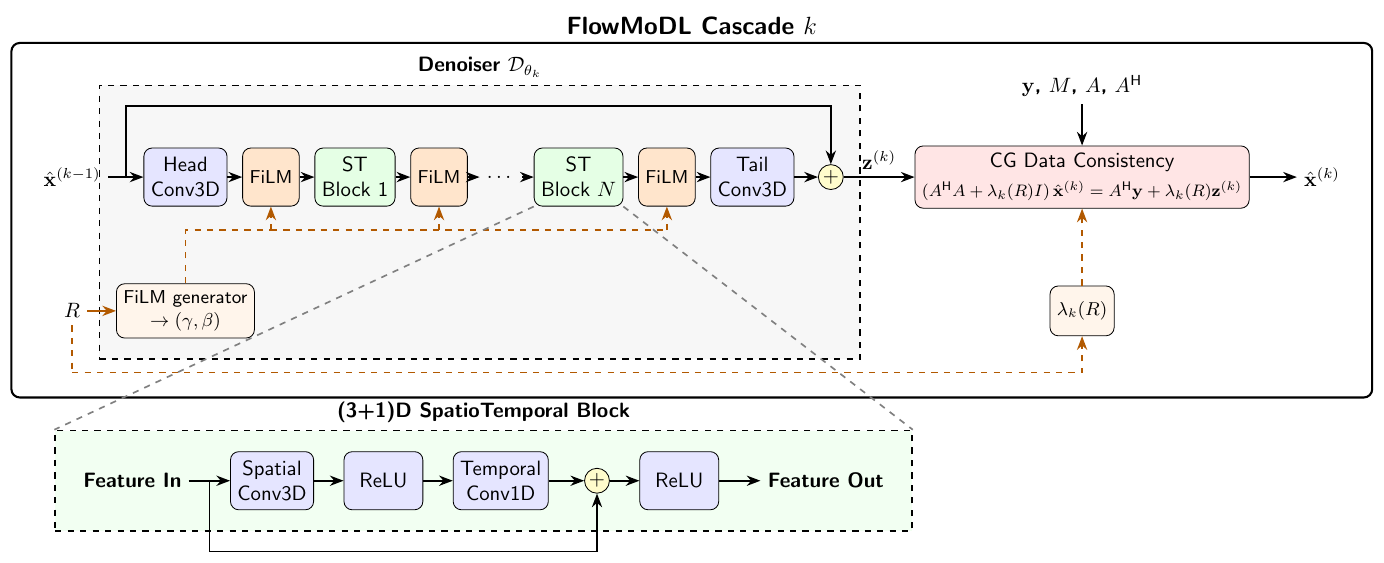}
    \caption{Detailed architecture of a single FlowMoDL cascade $k$ (top) and the internal structure of its factorized (3+1)D spatiotemporal blocks (bottom). The network maps the previous estimate $\hat{\mathbf{x}}^{(k-1)}$ to an intermediate denoised target $\mathbf{z}^{(k)}$ using a learned prior $\mathcal{D}_{\theta_k}$ equipped with a global residual connection. To improve the handling of varying acceleration rates, $R$ is routed to both a FiLM generator within the denoiser and an MLP computing the data-consistency penalty weight $\lambda_k(R)$. The cascade concludes by solving the linear data-consistency equations via conjugate-gradient iterations.} \label{fig:flowmodl_block}
\end{figure}
We propose FlowMoDL, a hybrid-unrolled neural network based on the \textit{Model-based reconstruction using Deep Learned priors} (MoDL) paradigm~\cite{aggarwal2018modl}. As shown in \Cref{fig:flowmodl_block}, the reconstruction is unrolled into $K$ cascades. 
The first cascade is initialized with the adjoint reconstruction $\hat{\mathbf{x}}^{(0)}=A^{\mathsf H}\mathbf{y}$.\\
In each cascade $k\in\{1,\dots,K\}$, the current estimate $\hat{\mathbf{x}}^{(k-1)}$ is mapped to a denoised intermediate $\mathbf{z}^{(k)}$ by the learned denoiser $\mathcal{D}_{\theta_k}$.
Unlike the original MoDL formulation, we condition the denoiser on the scalar acceleration factor $R$ via feature-wise linear modulation (FiLM)~\cite{perez2018film}:
\begin{equation}
    \mathbf{z}^{(k)} = \mathcal{D}_{\theta_k}(\hat{\mathbf{x}}^{(k-1)}, R)
\end{equation}
Holding $\mathbf{z}^{(k)}$ fixed, the subsequent data-consistency update obtains the cascade output by solving
\begin{equation}
    \left(A^{\mathsf H}A+\lambda_k(R)I\right)
    \hat{\mathbf{x}}^{(k)}
    =
    A^{\mathsf H}\mathbf{y}
    +
    \lambda_k(R)\mathbf{z}^{(k)}.
    \label{eq:flowmodl_cascade}
\end{equation}
This linear system is approximately solved using $J$ iterations of the conjugate gradient (CG) method. To reduce the computational cost of each CG iteration, we transform the fully sampled readout axis into image space before the unrolled cascades, exploiting the undersampling mask being constant along this axis~\cite{cmrx4dflow2026data}. The resulting normal operator decomposes into independent
2D problems indexed by readout position, allowing each CG iteration to be performed using 2D FFTs while remaining equivalent to the full 3D data-consistency formulation.\\
\textbf{Regularizer:}
To enable the network to compensate for higher acceleration factors compared to the original MoDL~\cite{aggarwal2018modl}, each cascade utilizes an independent denoiser network, $\mathcal{D}_{\theta_k}$. The denoiser processes the image by concatenating the real and imaginary parts of all velocity encodings along the channel axis, preserving cross-encoding phase relationships across layers. As detailed in the bottom panel of \Cref{fig:flowmodl_block}, the backbone comprises $N$ residual spatiotemporal blocks. Instead of separate hyperplane convolutions~\cite{vishnevskiy2020flowvn}, each block employs a (3+1)D factorization~\cite{tran2018closer}: a spatial 3D convolution per cardiac phase is sequentially followed by a ReLU activation and a temporal 1D convolution per voxel. This temporal convolution aims to capture temporal flow dynamics without requiring the recurrent hidden states used in FlowMRI-Net, which substantially slows down training~\cite{jacobs2025flowmrinet}.
As shown by the global skip connection around the denoiser in \Cref{fig:flowmodl_block}, the zero-initialized final convolution is added to the denoiser input. Consequently, each denoiser initially returns its input, providing a stable initialization of the learned regularization pathway before the subsequent CG data-consistency update.\\
\textbf{Data consistency:}
The update in \Cref{eq:flowmodl_cascade} is approximated using \(J\) \linebreak conjugate-gradient iterations. Since we enforce \(\lambda_k(R)>0\), the system operator is Hermitian positive definite, ensuring that the corresponding linear system has a unique solution that is approximated by the unrolled CG iterations. Their differentiable implementation allows gradients to propagate through the data-consistency update. \\
\textbf{Acceleration Conditioning:} 
To handle varying acceleration factors with a single trained model, we introduce a dual-pathway conditioning scheme based on the log-scaled and normalized acceleration factor $R$. First, $R$ is mapped by a FiLM generator to per-channel scales and shifts ($\gamma, \beta$), which are injected into the feature maps immediately following the denoiser stem and each subsequent residual block. Simultaneously, a separate Multi-Layer Perceptron (MLP) uses $R$ to predict a per-sample offset for each cascade's regularization weight, $\lambda_k(R)$. This joint modulation allows both the denoiser features and the coupling between the reconstruction and denoised estimate to adapt to the acceleration factor. Both conditioning networks feature zero-initialized output layers, ensuring training begins entirely unconditioned.\\
\textbf{Optimization Objective}
Approaches focused on flow reconstruction like \cite{vishnevskiy2020flowvn,jacobs2025flowmrinet} optimize their networks using only a weighted per-layer L1-loss. While the L1-loss implicitly optimizes phase, in practice it heavily biases the network toward magnitude-dominant reconstructions. FlowMoDL includes three additional loss terms specifically penalizing phase misalignment: a velocity error loss, a relative error loss, and an angular error loss.
Given the chosen velocity encoding parameters $\mathbf{v}_{\mathrm{enc}}$, the ground truth velocity vector field $\mathbf{v}$ and predicted velocity vector field $\hat{\mathbf{v}}^{(k)}$ are calculated from the complex-valued directional encodings (indices 1 to 3) and the reference encoding (index 0) as $\mathbf{v} \;=\; 1/\pi\, \mathbf{v}_{\mathrm{enc}} \odot \angle\!\big(\boldsymbol{x}_{1:3}\,\overline{\mathbf{x}}_0\big)$.
Given the spatial segmentation mask $s \in \{0,1\}^{N_z\times N_y\times N_x}$ delineating the ROI, which is broadcast across the temporal and velocity dimensions, the unnormalized per-cascade image and velocity L1 terms over the spatiotemporal domain $\Omega$ are defined as:
\begin{equation}
    \begin{aligned}     
    \mathcal{L}^{(k)}_{\mathrm{img}} = \sum_{\Omega} s &\odot \big\Vert{}\hat{\mathbf{x}}^{(k)}-\mathbf{x}\big\Vert{}_1, \quad     
    \mathcal{L}^{(k)}_{\mathrm{vel}} = \sum_{\Omega} s \odot \big\Vert{}\hat{\mathbf{v}}^{(k)}-\mathbf{v}\big\Vert{}_1, \\     
    \mathcal{L}^{(k)}_{\mathrm{rel}} &= \sqrt{\frac{\sum_{\Omega} s \odot \big(\Vert{}\mathbf{v}\Vert{}_2-\Vert{}\hat{\mathbf{v}}^{(k)}\Vert{}_2\big)^2\, }{\sum_{\Omega} s \odot \Vert{}\mathbf{v}\Vert{}_2^2\, +\varepsilon_{\mathrm{rel}}}} \end{aligned}
\end{equation}
To penalize directional errors independently of magnitude, we introduce a voxel-wise angular cosine-distance surrogate: \begin{equation}
\begin{aligned}
\mathcal{L}^{(k)}_{\mathrm{ang}}
&=
\sum_{\Omega} s \odot \left(1-\cos\theta\right), \quad
\cos\theta = \mathrm{clip}\!\left( \frac{\hat{\mathbf{v}}^{(k)} \cdot \mathbf{v}}{\Vert{}\hat{\mathbf{v}}^{(k)}\Vert{}_2\,\Vert{}\mathbf{v}\Vert{}_2+\varepsilon_{\mathrm{ang}}}, \,-1,\,1\right).
\end{aligned}
\end{equation}
The total loss is a deep-supervision composite combining these metrics:
\begin{equation}
  \mathcal{L} \;=\; \sum_{k=1}^{K} w_k\,\Bigl[\, \mathcal{L}_{\mathrm{img}}^{(k)} + \omega_{\mathrm{vel}}\,\alpha_{\mathrm{v}}(e)\,\mathcal{L}_{\mathrm{vel}}^{(k)} + \omega_{\mathrm{rel}}\,\alpha_{\mathrm{v}}(e)\,\mathcal{L}_{\mathrm{rel}}^{(k)} + \omega_{\mathrm{ang}}\,\alpha_{\theta}(e)\,\mathcal{L}_{\mathrm{ang}}^{(k)} \Bigr], 
\end{equation}
where $k$ indexes the cascade output, $e$ is the current epoch, and $w_k = \exp\bigl(-\tau\,(K - k)\bigr)$ is an exponentially-decaying weight that concentrates gradients on later cascades as $\tau$ increases.
By including velocity and phase explicitly in the loss, the network optimizes directly for the targeted clinical metrics. To improve stability, the scalar multipliers $\alpha_{\mathrm{v}}(e)$ and $\alpha_{\theta}(e)$ apply a curriculum schedule~\cite{bengio2009curriculum}: the magnitude is optimized first, and the phase terms are ramped up subsequently.\\
\textbf{Training Procedure \& Hyperparameters}
FlowMoDL is trained end-to-end using AdamW~\cite{loshchilov2017decoupled} with gradient clipping and a cosine-annealing learning rate schedule ($10^{-3}$ to $10^{-6}$). SENSE operations and the CG solver compute in single precision, while the convolutional backbone uses mixed precision. Gradient checkpointing is used to fit training on a \qty{48}{\giga\byte} GPU.
The architecture unrolls $K=8$ untied cascades, each performing $J=10$ CG iterations with an independent regularization weight (initialized to $\lambda_k(R) = 0.05$). The denoiser uses $64$ channels, $N=4$ residual blocks, a $3{\times}3{\times}3$ spatial kernel, and a length-$5$ temporal kernel. Loss weights are configured to $\omega_{\mathrm{vel}} = \omega_{\mathrm{rel}} = \omega_{\mathrm{ang}} = 0.1$. The curriculum multipliers $\alpha_{\mathrm{v}}(e)$ and $\alpha_{\theta}(e)$ increase linearly over $8$ epochs, starting at epochs $4$ and $12$, respectively. To prevent overfitting, training relies on random spatiotemporal crops of $16 \times 64 \times 64$ spatial voxels across $6$ cardiac frames.

\section{Experiments}
\subsection{Data}
We utilize the multi-center, multi-vendor CMRx4DFlow challenge dataset~\cite{cmrx4dflow2026data}, partitioning the 138 original training cases into custom training (96 cases), validation (21 cases), and test (21 cases) sets preserving the original center and scanner architecture distributions of the original cohort. During training, kt-Gaussian undersampling masks are generated randomly on the fly to match a randomly selected target acceleration factor. For validation and testing, we evaluate the models using undersampling masks pre-generated according to the same protocol at acceleration factors ranging from $10\times$ to $50\times$.

\subsection{Main Experiments}
\begin{figure}[tb]
    \centering
    \begin{subfigure}[b]{0.4\linewidth}
        \centering
        \includegraphics[width=\linewidth]{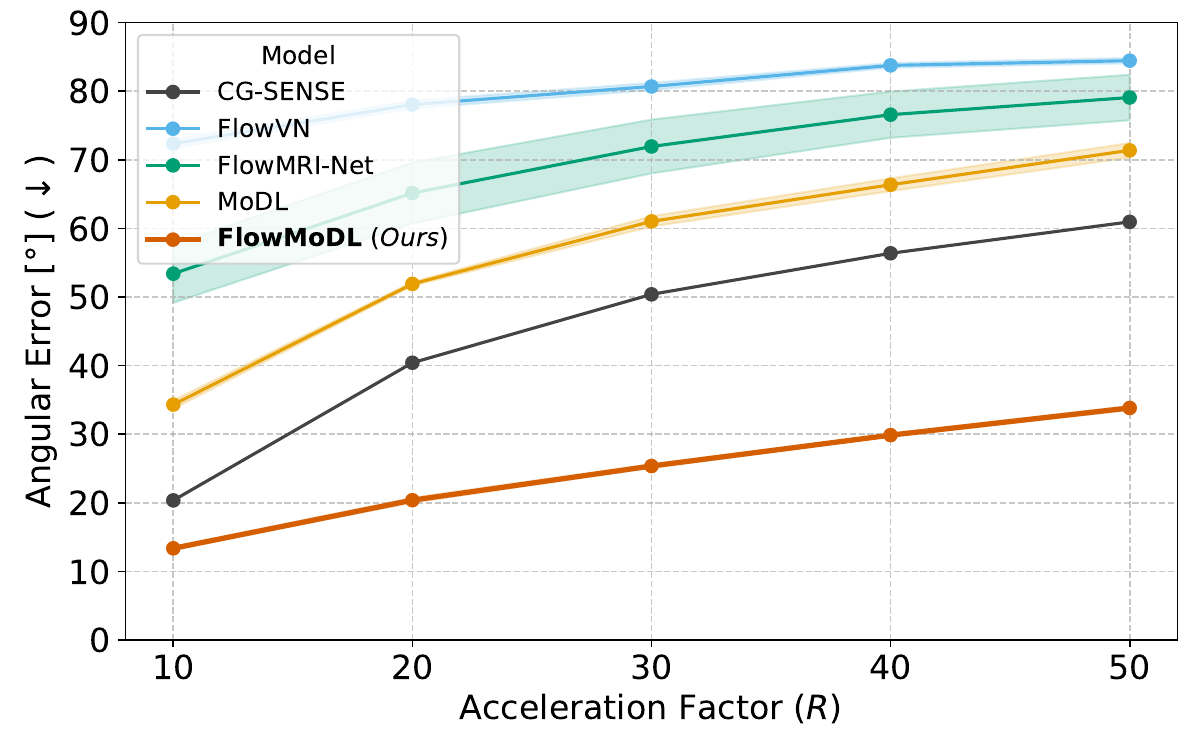}
        \caption{Angular Error [°] ($\downarrow$)}
        \label{fig:angerr}
    \end{subfigure}\hfill
    \begin{subfigure}[b]{0.4\linewidth}
        \centering
        \includegraphics[width=\linewidth]{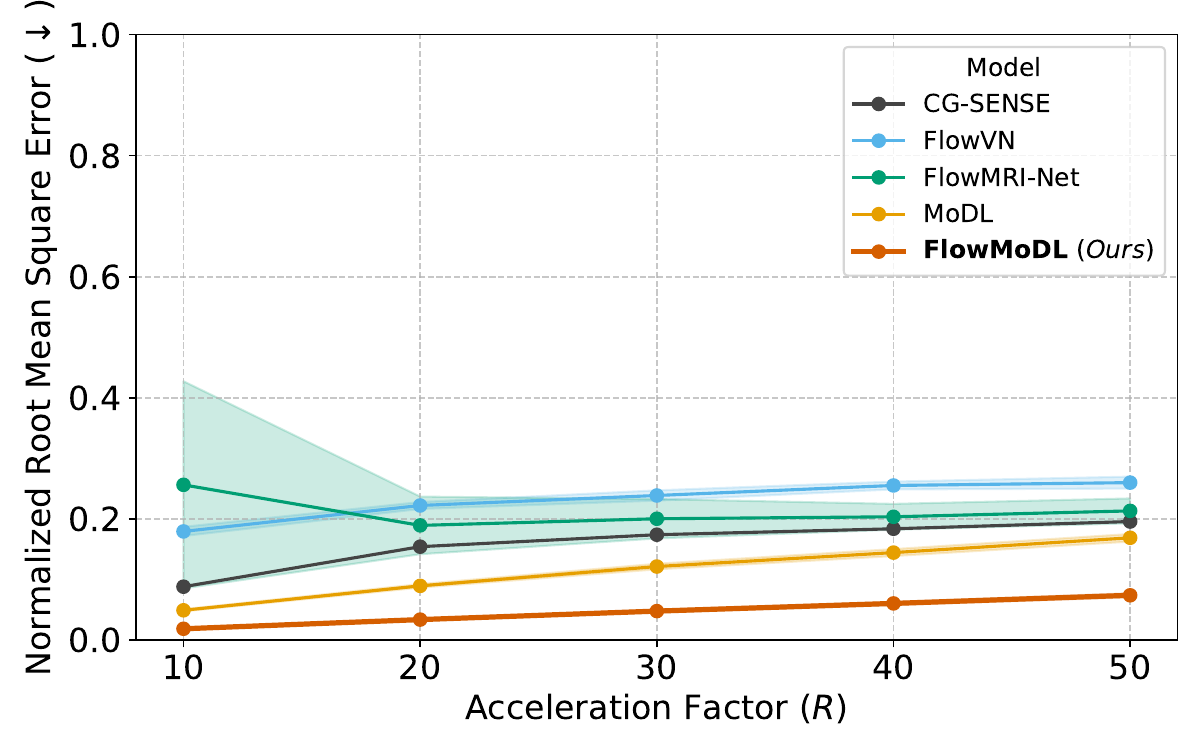}
        \caption{nRMSE ($\downarrow$)}
        \label{fig:nrmse}
    \end{subfigure}
    \begin{subfigure}[b]{0.4\linewidth}
        \centering
        \includegraphics[width=\linewidth]{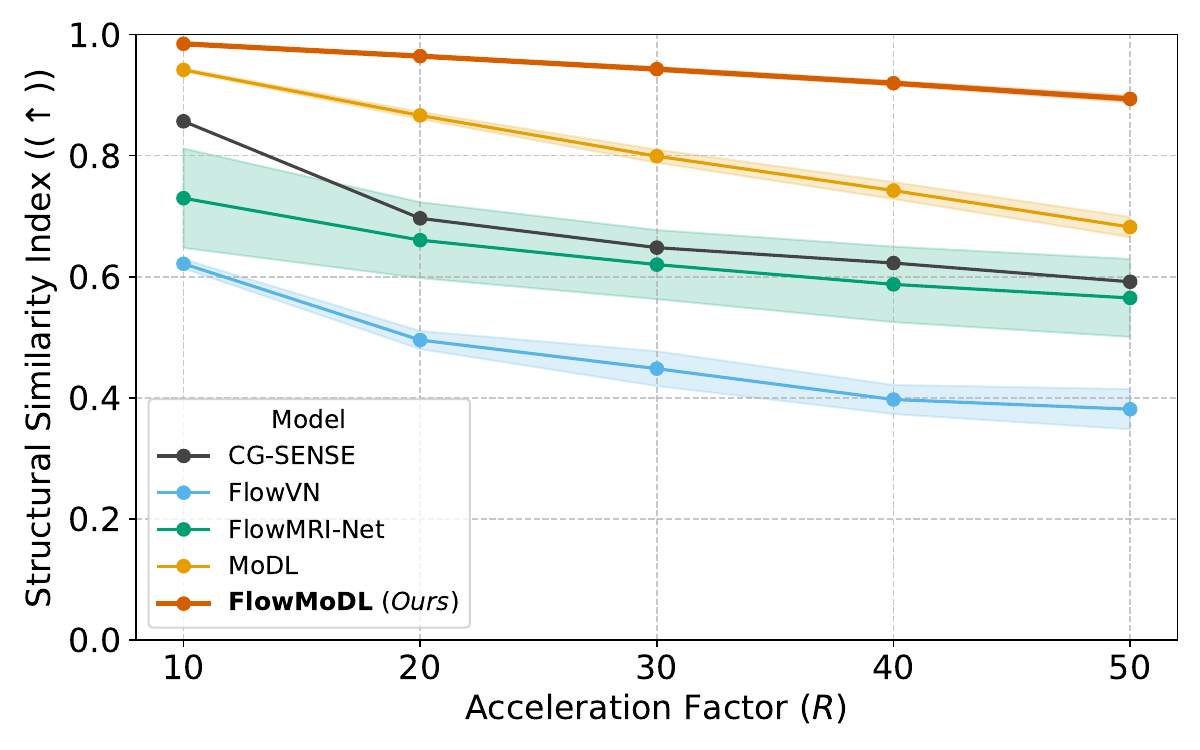}
        \caption{SSIM ($\uparrow$)}
        \label{fig:ssim}
    \end{subfigure}\hfill
    \begin{subfigure}[b]{0.4\linewidth}
        \centering
        \includegraphics[width=\linewidth]{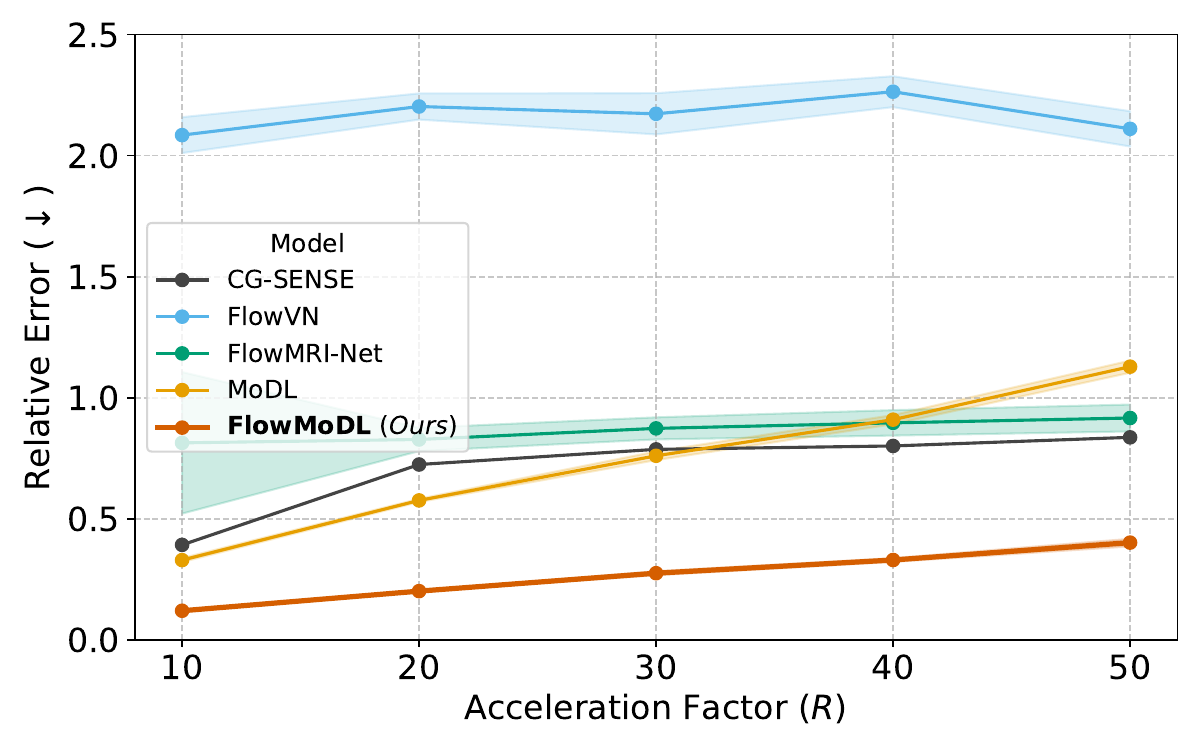}
        \caption{Relative Error ($\downarrow$)}
        \label{fig:relerr}
    \end{subfigure}
    \caption{Metrics across different $R$'s. For deep learning methods, solid lines denote the mean and shaded areas represent standard deviation ($n=5$ seeds).}
    \label{fig:acceleration_factor_performance}
\end{figure}

\begin{figure}[htb]\centering
    \includegraphics[width=0.9\linewidth]{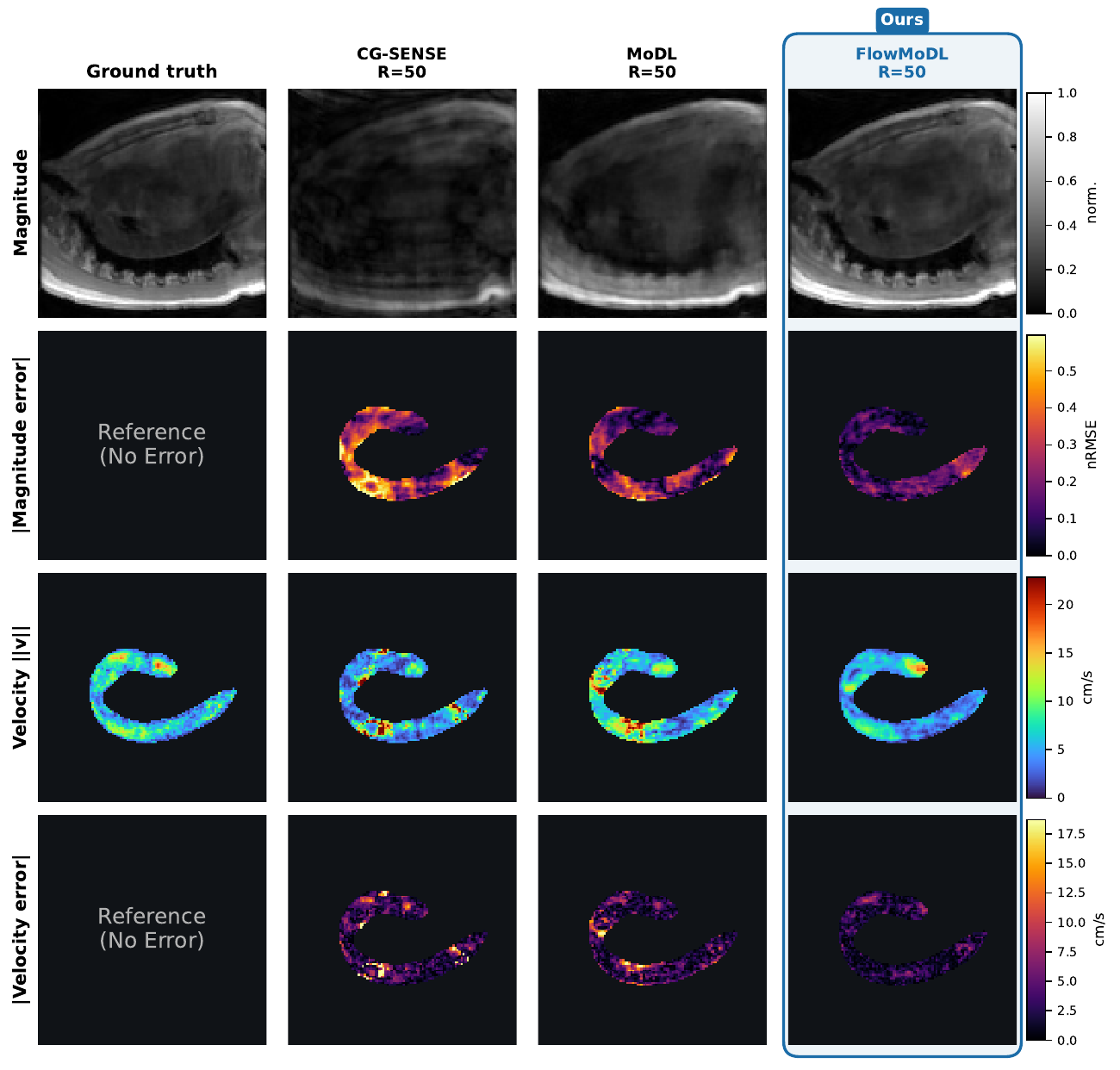}
    \caption{Representative qualitative results at $R=50$ comparing CG-SENSE, MoDL, and the proposed FlowMoDL with the fully sampled reference. FlowMoDL reconstructs the volume with the lowest magnitude and velocity errors.} \label{fig:recon_vis}
\end{figure}

\begin{table*}[tb]
    \centering
    \caption{Mean performance of reconstruction models on the unseen test set averaged across all acceleration factors ($R \in \{10, 20, 30, 40, 50\}$). The best results are highlighted in \textbf{bold}, and the second-best results are \underline{underlined}.}
    \label{tab:mri_mean_results}
    \small
    \resizebox{\textwidth}{!}{ 
    \begin{tabular}{l|rrrr r}
        \toprule
        \textbf{Model} & \textbf{nRMSE} ($\downarrow$) & \textbf{SSIM} ($\uparrow$) & \textbf{RelErr} ($\downarrow$) & \textbf{AngErr} ($\downarrow$) \\
        \midrule
        CG-SENSE~\cite{pruessmann2001advances} & 0.1592 & 0.6831 & \underline{0.7087} & \underline{45.6948}\\
        FlowVN~\cite{vishnevskiy2020flowvn} & 0.2311 $\pm$ 0.0061 & 0.4687 $\pm$ 0.0200 & 2.1668 $\pm$ 0.0648 & 79.8481 $\pm$ 0.4238 \\
        FlowMRI-Net~\cite{jacobs2025flowmrinet} & 0.2126 $\pm$ 0.0540 & 0.6325 $\pm$ 0.0602 & 0.8660 $\pm$ 0.0741 & 69.2204 $\pm$ 3.7776\\
        MoDL~\cite{aggarwal2018modl} & \underline{0.1147} $\pm$ 0.0035 & \underline{0.8062} $\pm$ 0.0095 & 0.7412 $\pm$ 0.0126 & 56.9912 $\pm$ 0.6201\\
        \textbf{FlowMoDL (Ours)} & \textbf{0.0469} $\pm$ 0.0017  & \textbf{0.9409} $\pm$ 0.0026  & \textbf{0.2664} $\pm$ 0.0065  & \textbf{24.5639} $\pm$ 0.2458  \\
        \bottomrule
    \end{tabular}
    }
\end{table*}
We evaluate FlowMoDL against classical \textit{CG-SENSE}~\cite{pruessmann2001advances}, the standard \textit{MoDL} architecture~\cite{aggarwal2018modl}, \textit{FlowVN}~\cite{vishnevskiy2020flowvn} and \textit{FlowMRI-Net}~\cite{jacobs2025flowmrinet}. To ensure a fair comparison, all learned models were trained for an equivalent number of gradient steps. This corresponds to 50 epochs using standard random spatiotemporal cropping, avoiding the highly correlated, sliding-window epoch definitions originally proposed for FlowVN and FlowMRI-Net. Performance is measured via magnitude structural similarity (SSIM)~\cite{wang2004ssim}, magnitude nRMSE, velocity relative error (RelErr) in the aorta, and velocity angular error (AngErr)~\cite{vishnevskiy2020flowvn}.\\
\Cref{tab:mri_mean_results} and \Cref{fig:acceleration_factor_performance} summarize the quantitative results on the unseen test set, averaged across five random seeds. FlowMoDL strictly outperforms all baselines, including MoDL, across all acceleration factors ($R \in \{10, \dots, 50\}$). While MoDL provides a robust baseline for basic structural reconstruction, FlowMoDL's explicit phase-aware objective and spatiotemporal formulation yield substantially lower angular and relative errors, while still outperforming MoDL on nRMSE and SSIM. Notably, FlowVN and FlowMRI-Net degrade significantly under this normalized gradient-step budget, underperforming classical CG-SENSE.
Qualitative improvements are evident in \Cref{fig:recon_vis}. At a high acceleration rate ($R=50$), both CG-SENSE and MoDL struggle with high-frequency spatial details and suffer from velocity artifacts. In contrast, FlowMoDL's learned prior recovers sharp spatial structures and smooth velocity transitions. 
Beyond spatial fidelity, FlowMoDL preserves temporal coherence. Since blood flow $\mathbf{v}(t)$ is naturally continuous while residual undersampling noise is temporally uncorrelated, we quantify temporal coherence using the lag-one autocorrelation $\rho_1$ and the fraction of temporal power above half the Nyquist frequency $\eta$~\cite{box2015time}. At $R=50$, CG-SENSE ($\rho_1 \approx 0.00$, $\eta \approx 0.53$) and MoDL ($\rho_1 \approx -0.05$, $\eta \approx 0.55$) yield metrics indicative of temporally white noise artifacts. Conversely, FlowMoDL ($\rho_1 \approx 0.72$, $\eta \approx 0.07$) closely matches the ground truth ($\rho_1 \approx 0.71$, $\eta \approx 0.09$), successfully reconstructing temporally coherent velocity fields with only marginal over-smoothing.

\section{Conclusion}
In this work, we presented FlowMoDL, an efficient model-based unrolled network tailored for highly accelerated 4D flow MRI reconstruction. By integrating a (3+1)D spatiotemporal learned prior with conjugate-gradient data-consistency and dynamic conditioning for the acceleration factor, FlowMoDL effectively recovers both anatomical structures and complex flow dynamics using a single trained model. Crucially, optimizing directly for phase alignment via a specialized curriculum-based composite loss, incorporating velocity, relative, and angular error terms, prevents the magnitude-bias common in generic MRI reconstruction tasks. Evaluated on the multi-center CMRx4DFlow dataset, FlowMoDL strictly outperforms classical CG-SENSE, standard MoDL, and specialized deep learning architectures (FlowVN and FlowMRI-Net) across acceleration factors from $10\times$ to $50\times$. Furthermore, we demonstrated that FlowMoDL exhibits superior training efficiency compared to existing flow-specific networks under a normalized gradient-step budget, consistently yielding sharp magnitude reconstructions and physiologically accurate, temporally coherent velocity fields even at extreme undersampling rates.

\begin{credits}
\subsubsection{\ackname} 
This work was supported by DFG Heisenberg (502024488), ERC StG EARTHWORM (101165242), ERC Proof-of-concept grant SYNCWORM\linebreak (101293293) and CAIMed - Lower Saxony Center for Artificial Intelligence and Causal Methods in Medicine (ZN4257).

\subsubsection{\discintname}
The authors have no competing interests to declare that are
relevant to the content of this article.
\end{credits}

\newpage
%
%
\bibliographystyle{splncs04}
\bibliography{references}
\end{document}